\documentclass[sigconf]{acmart}
\AtBeginDocument{%
  \providecommand\BibTeX{{%
    \normalfont B\kern-0.5em{\scshape i\kern-0.25em b}\kern-0.8em\TeX}}}

\setcopyright{acmlicensed}
\copyrightyear{2024}
\acmYear{2024}
\setcopyright{acmlicensed}\acmConference[MMASIA '24]{ACM Multimedia Asia}{December 3--6, 2024}{Auckland, New Zealand}
\acmBooktitle{ACM Multimedia Asia (MMASIA '24), December 3--6, 2024, Auckland, New Zealand}
\acmDOI{10.1145/3696409.3700223}
\acmISBN{979-8-4007-1273-9/24/12}

\usepackage{multirow}
\begin{document}

%%
%% The "title" command has an optional parameter,
%% allowing the author to define a "short title" to be used in page headers.
\title{Exploring Annotation-Free Image Captioning with Retrieval-Augmented Pseudo Sentence Generation}

%%
%% The "author" command and its associated commands are used to define
%% the authors and their affiliations.
%% Of note is the shared affiliation of the first two authors, and the
%% "authornote" and "authornotemark" commands
%% used to denote shared contribution to the research.

\author{Zhiyuan Li}
\affiliation{%
  \institution{University of Sydney}
  % \streetaddress{1 Th{\o}rv{\"a}ld Circle}
  \city{Sydney}
  \country{Australia}}
\email{zhli0736@uni.sydney.edu.au}

\author{Dongnan Liu}
\affiliation{%
  \institution{University of Sydney}
  % \streetaddress{1 Th{\o}rv{\"a}ld Circle}
  \city{Sydney}
  \country{Australia}}
\email{dongnan.liu@sydney.edu.au}

\author{Heng Wang}
\affiliation{%
  \institution{University of Sydney}
  % \streetaddress{1 Th{\o}rv{\"a}ld Circle}
  \city{Sydney}
  \country{Australia}}
\email{hwan9147@uni.sydney.edu.au}

\author{Chaoyi Zhang}
\affiliation{%
  \institution{University of Sydney}
  % \streetaddress{1 Th{\o}rv{\"a}ld Circle}
  \city{Sydney}
  \country{Australia}}
\email{czha5168@uni.sydney.edu.au}

\author{Weidong Cai}
\affiliation{%
  \institution{University of Sydney}
  % \streetaddress{1 Th{\o}rv{\"a}ld Circle}
  \city{Sydney}
  \country{Australia}}
\email{tom.cai@sydney.edu.au}
%%
%% By default, the full list of authors will be used in the page
%% headers. Often, this list is too long, and will overlap
%% other information printed in the page headers. This command allows
%% the author to define a more concise list
%% of authors' names for this purpose.
\renewcommand{\shortauthors}{Zhiyuan Li, et al.}

%%
%% The abstract is a short summary of the work to be presented in the
%% article.
\begin{abstract}
  Recently, training an image captioner without annotated image-sentence pairs has gained traction. Previous methods have faced limitations due to either using mismatched corpora for inaccurate pseudo annotations or relying on resource-intensive pre-training. To alleviate these challenges, we propose a new strategy where the prior knowledge from large pre-trained models (LPMs) is distilled and leveraged as supervision, and a retrieval process is integrated to further reinforce its effectiveness. Specifically, we introduce \textbf{R}etrieval-\textbf{a}ugmented \textbf{P}seudo \textbf{S}entence \textbf{G}eneration (RaPSG), which can efficiently retrieve highly relevant short region descriptions from the mismatching corpora and use them to generate a variety of high-quality pseudo sentences via LPMs. Additionally, we introduce a fluency filter and a CLIP guidance objective to enhance contrastive information learning. Experimental results indicate that our method outperforms SOTA captioning models across various settings including zero-shot, unsupervised, semi-supervised, and cross-domain scenarios. Code is available at: https://github.com/Zhiyuan-Li-John/RaPSG.
\end{abstract}

%%
%% The code below is generated by the tool at http://dl.acm.org/ccs.cfm.
%% Please copy and paste the code instead of the example below.
%%
\begin{CCSXML}
<ccs2012>
 <concept>
  <concept_id>00000000.0000000.0000000</concept_id>
  <concept_desc>Do Not Use This Code, Generate the Correct Terms for Your Paper</concept_desc>
  <concept_significance>500</concept_significance>
 </concept>
 <concept>
  <concept_id>00000000.00000000.00000000</concept_id>
  <concept_desc>Do Not Use This Code, Generate the Correct Terms for Your Paper</concept_desc>
  <concept_significance>300</concept_significance>
 </concept>
 <concept>
  <concept_id>00000000.00000000.00000000</concept_id>
  <concept_desc>Do Not Use This Code, Generate the Correct Terms for Your Paper</concept_desc>
  <concept_significance>100</concept_significance>
 </concept>
 <concept>
  <concept_id>00000000.00000000.00000000</concept_id>
  <concept_desc>Do Not Use This Code, Generate the Correct Terms for Your Paper</concept_desc>
  <concept_significance>100</concept_significance>
 </concept>
</ccs2012>
\end{CCSXML}

\ccsdesc[500]{Computing methodologies~Natural language generation}
\ccsdesc[500]{Computing methodologies~Image representations}
\ccsdesc[500]{Computing methodologies~Scene understanding}

%%
%% Keywords. The author(s) should pick words that accurately describe
%% the work being presented. Separate the keywords with commas.
\keywords{Image Captioning, Image-Text Retrieval, Large Pre-Trained Model}

%% A "teaser" image appears between the author and affiliation
%% information and the body of the document, and typically spans the
% %% page.
% \begin{teaserfigure}
%   \includegraphics[width=\textwidth]{sampleteaser}
%   \caption{Seattle Mariners at Spring Training, 2010.}
%   \Description{Enjoying the baseball game from the third-base
%   seats. Ichiro Suzuki preparing to bat.}
%   \label{fig:teaser}
% \end{teaserfigure}

% \received{20 February 2007}
% \received[revised]{12 March 2009}
% \received[accepted]{5 June 2009}

%%
%% This command processes the author and affiliation and title
%% information and builds the first part of the formatted document.
\maketitle

\section{Introduction}
\label{intro}
\begin{figure}[t]
  \centering
  %\fbox{\rule{0pt}{2in} \rule{0.9\linewidth}{0pt}}
   \includegraphics[width=0.99\linewidth]{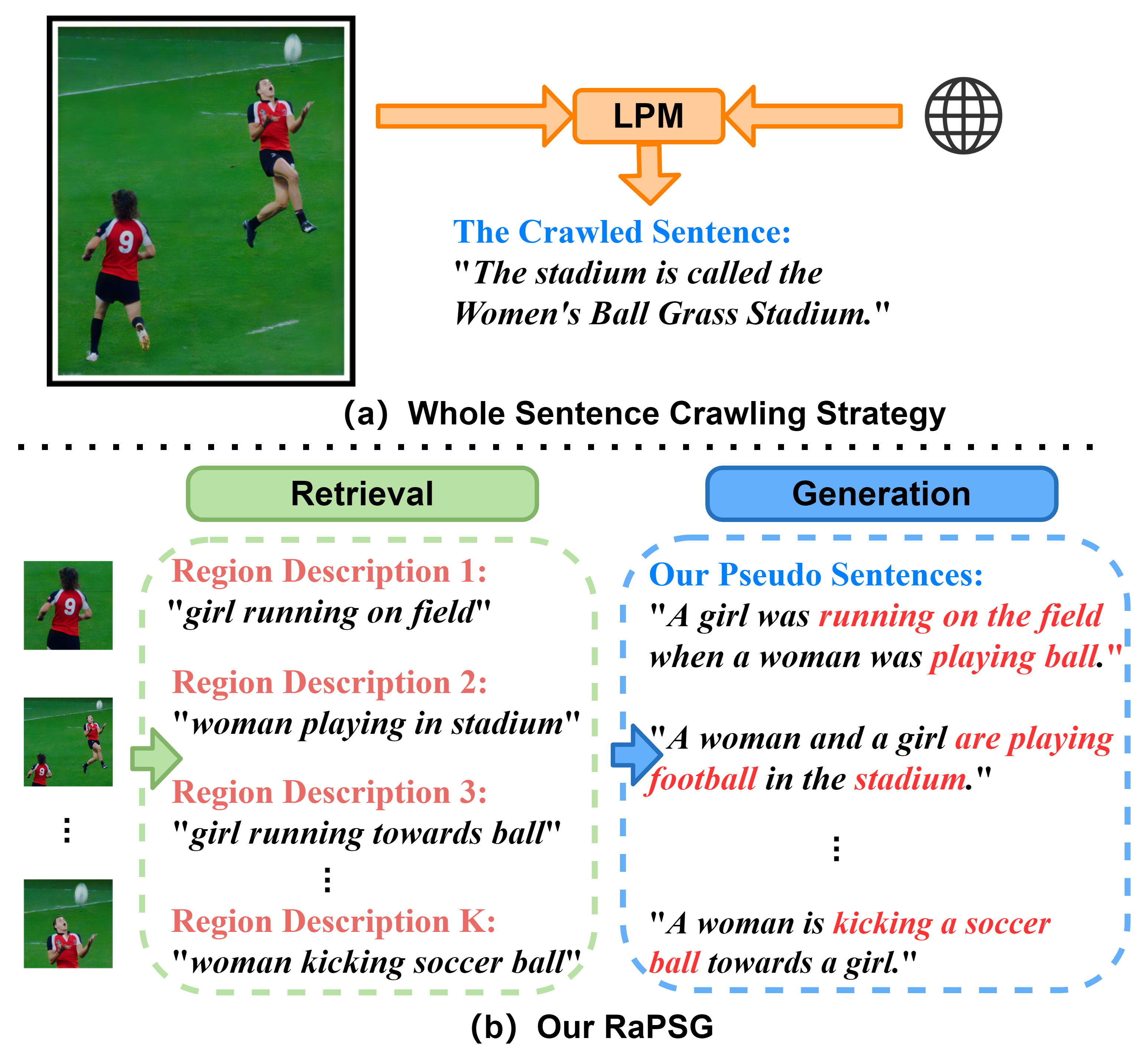}
   \caption{The comparison between whole sentence crawling strategy~\cite{kakaobrain2022coyo-700m} and our generation-based RaPSG method.}
   \label{fig1}
\end{figure}
Recent advancements in image captioning have been driven by Transformer-based models~\cite{cornia2020meshed,luo2021dual}. However, the reliance on high-quality human-annotated image-text pairs limits these fully supervised approaches, increasing interest in annotation-free alternatives, such as unsupervised and pre-training strategies. Unsupervised approaches~\cite{guo2020recurrent, zhou2021triple} align crawled sentences with target images as pseudo annotations, but face issues with sentence diversity~\cite{li2022blip} and content accuracy~\cite{honda2021removing}. Pre-training strategies show strong performance but require massive resources~\cite{wang2021simvlm} and are affected by noisy data from coarse LPM-led selection~\cite{kakaobrain2022coyo-700m}, as shown in Figure~\ref{fig1}(a), leading to poor sample efficiency~\cite{li2022blip}.

To alleviate these problems, recent methods transfer prior knowledge from frozen LPMs to vision-language (VL) tasks. Notable architectures like Flamingo~\cite{alayrac2022flamingo} and BLIP2~\cite{li2023blip} use trainable mapper modules to bridge LPMs with vision encoders, keeping LPMs frozen to reduce computational cost and avoid catastrophic forgetting. Similarly, LLaVA~\cite{liu2024visual} and MiniGPT4~\cite{zhu2023minigpt} employ projection layers to integrate visual encoders with language decoders, innovating through fine-tuning on multimodal instructions. However, despite these advancements, all these methods still rely on billions of external image-text pairs for "mapper" learning and remain susceptible to the challenge of the noisy image-text pairs problem.

In this paper, we propose an efficient Retrieval-augmented Pseudo Sentence Generation framework (RaPSG) that leverages prior knowledge from frozen LPMs as supervision by generating high-quality pseudo sentences without the need of external image-text pairs or instruction tuning for optimization. Specifically, a retrieval-based pipeline is designed to generate multiple sentences for each target image, as shown in Figure~\ref{fig1}(b). To address the challenge of noisy image-text pairs and improve the quality of generated pseudo sentences, we propose a refinement strategy based on a ranking of high-relevance region descriptions.  For each target image, we employ the pre-trained model CLIP~\cite{radford2021learning} to retrieve the top-$k$ most correlated region descriptions from the Visual Genome (VG) dataset~\cite{krishna2017visual} (We eliminate the overlapping parts between COCO and VG). Then, we further group region descriptions into multiple comprehensive and distinct long sentences using summarization LPMs, such as BART~\cite{lewis2019bart} and LLaMA~\cite{touvron2023llama}. After this, we then introduce a self-supervised framework to facilitate the retrieval-augmented captioner, using original images and generated pseudo sentences as supervision. Additionally, we design two mechanisms to enhance the plain pseudo-labeling strategy. Firstly, a fluency filter removes imperfect descriptions to mitigate the impact of noisy image-text pairs.  Second, a CLIP-based optimization strategy improves the model's comprehension of image-text pairs, offsetting the lack of external image-text pairs. 

To demonstrate the capability of our RaPSG approach, we evaluate its performance on the MSCOCO~\cite{chen2015microsoft} and Flickr30k~\cite{plummer2015flickr30k} benchmarks across various settings. The results show that our method outperforms the SOTA captioner Flamingo3B with fewer trainable parameters and consistently surpasses other models in pre-training, unsupervised, weakly supervised, and unpaired settings. This highlights its effectiveness and efficiency. Additionally, we validate its robustness in semi-supervised and cross-domain settings, where our model also achieves SOTA performance, underscoring its versatility.

Our contributions are summarized as four folds: 
\begin{itemize}
    \item (1) We propose an inference-only approach that distils knowledge from frozen LLMs by retrieving highly relevant region descriptions and generating a variety of distinct pseudo sentences for each target image. 
    \item (2) A fluency filter and CLIP guidance are further introduced to strengthen the retrieval-augmented learning of the captioner for better prediction.
    \item (3) Experimental findings reveal that our approach surpasses current SOTA captioning models in a range of scenarios, including zero-shot, unsupervised, semi-supervised and cross-domain settings. 
    \item   (4) In our experiments, we also find that using high-quality generated pseudo sentences is more beneficial for captioner training than retrieved complete sentences, even if they are unpaired and sourced directly from the original dataset.
\end{itemize}

\section{Related Work}
\subsection{Large Pre-trained Models for Image Captioning}
In recent years, the appearance of a series of high-performance LPMs such as ViT~\cite{dosovitskiy2020image}, GPT-2~\cite{radford2019language}, and CLIP~\cite{radford2021learning} has widely extended the possibility of getting prior knowledge. \citet{kuo2022beyond} used CLIP to mine missing attributes and relationships as auxiliary inputs in a fully supervised captioning task. \citet{cho2022fine} used CLIP to build a CLIP score replacing the traditional cross-entropy loss, which can avoid references in strength learning of captioning tasks. Additionally, some works started to explore leveraging from the frozen LPMs. Flamingo~\cite{alayrac2022flamingo} builds a trainable architecture that bridges the vision encoder and the large language model, efficiently accepting arbitrarily interleaved visual data, and generating text in an open-ended manner. BLIP-2~\cite{li2023blip} bridges the modality gap with a lightweight querying Transformer and is more efficient in the pre-training strategy. However, all these methods still need pre-training on large-scale datasets for model optimization. 

\begin{figure}[t]
  \centering
  %\fbox{\rule{0pt}{2in} \rule{0.9\linewidth}{0pt}}
   \includegraphics[width=1.0\linewidth]{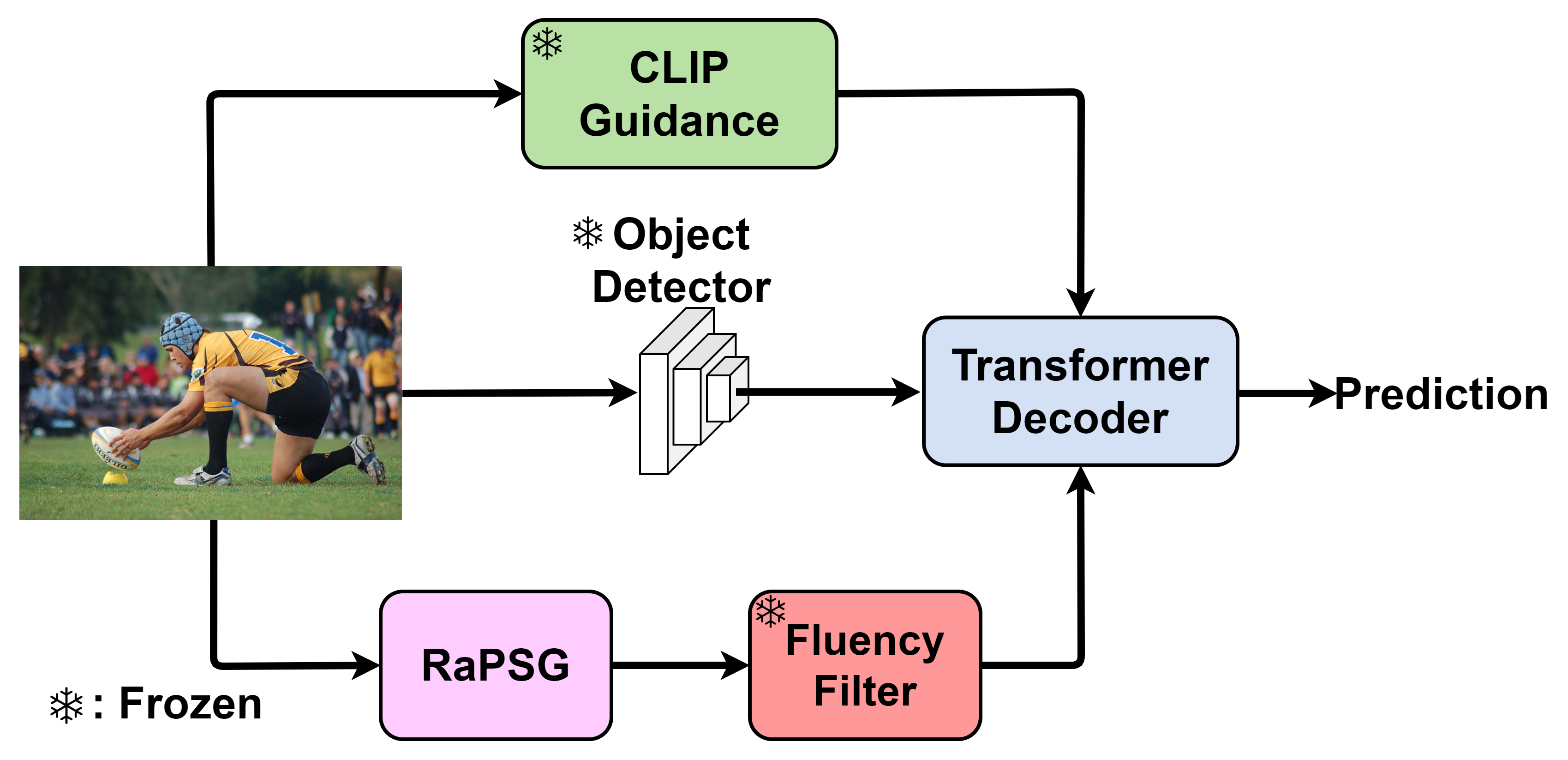}
   \vspace{0.0cm}
   \caption{The overview of our proposed framework. It is structured around three core components: RaPSG, fluency filter, and CLIP guidance. Notably, both the fluency filter and CLIP guidance modules are designed to be frozen, eliminating the need for further parameter training.}
   \label{fig2}
   \vspace{-0.1cm}
\end{figure}

\begin{figure*}[t]
  \centering
  %\fbox{\rule{0pt}{2in} \rule{0.9\linewidth}{0pt}}
   \includegraphics[width=0.88\linewidth]{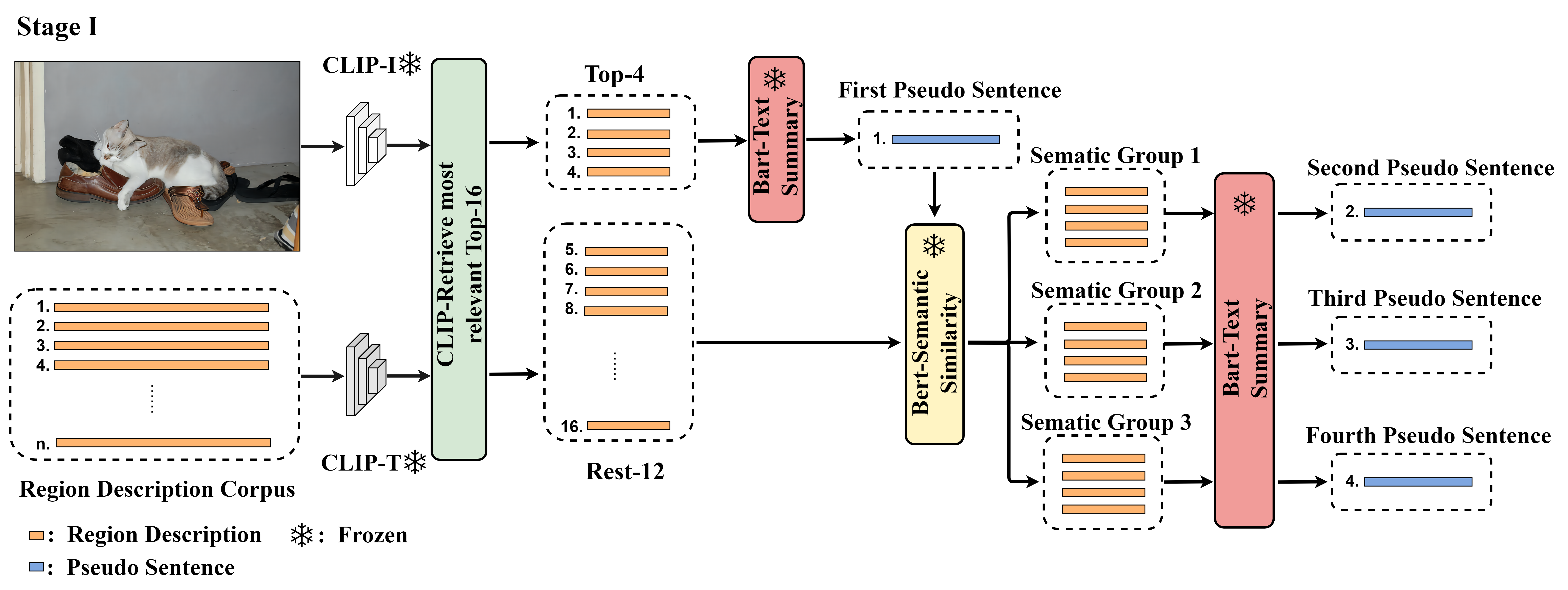}
   \vspace{-0.0cm}
   \caption{The Stage-I of RaPSG framework. Firstly, we retrieve top-$k$ region descriptions from VG~\cite{krishna2017visual} according to their matching scores computed by CLIP~\cite{radford2021learning} model. Then, we use Sent-BERT~\cite{reimers2019sentence} model to divide them into four groups by their semantic similarity. Finally, BART~\cite{lewis2019bart} model is used to summarize the grouped descriptions for four pseudo sentences.}
   \label{fig3}
   \vspace{-0.0cm}
\end{figure*}

\subsection{Retrieval-augmented Models with LPMs}
Retrieval-augmented methods have been widely applied in VL tasks in recent years. In visual question answering, retrieving the outside knowledge for question answering has become the new trend~\cite{lin2022retrieval, li2023multi, li2024enhancing, li2024multimodal}. In text-to-image generation, \citet{chen2022re} propose a generative model that uses retrieved information to produce high-fidelity images for uncommon entities. Currently, few works apply a retrieval-augmented idea with LPMs for image captioning. \citet{zhu2023prompt} use CLIP to extract the semantic prompt for more accurate caption prediction under the adversarial learning framework. Re-ViLM~\cite{yang2023re} builds upon the Flamingo but supports using CLIP to retrieve relevant knowledge from the external database. Compared with their methods, our approach gets knowledge from high-quality generated pseudo sentences and is more data-efficient, which avoids using unpaired human annotation~\cite{zhu2023prompt} or large-scale image-text corpus for pre-training~\cite{yang2023re}.

\section{Method}
In this section, we introduce our proposed framework RaPSG, whose overview is shown in Figure~\ref{fig2}. The retrieval-augmented pseudo sentence efficient generation module is proposed to learn knowledge from the LPMs (Section~\ref{pseudo}). To reduce the appearance of unnatural pseudo sentences, we innovatively design a fluency filter (Section~\ref{fluency}). Finally, the self-supervised training with generated pseudo image-text pairs is guided by a CLIP-based loss to improve the prediction accuracy (Section~\ref{contrastive}).

\subsection{Retrieval-Augmented PSG Module}
\label{pseudo}
To address the absence of human annotation, we propose RaPSG, a two-stage retrieval-augmented pseudo sentence generation method. It leverages the prior knowledge in LPMs to generate high-quality pseudo sentences for effective training supervision. Specifically, our method is based on the text processing capabilities from different aspects of LPMs including region-level matching with CLIP, global-level summarization through BART, and LLaMA for further enhancement. Stage-I transforms region-level information into global-level sentences to establish context, while Stage-II distills and refines these sentences with detailed content. This approach ensures high-quality pseudo sentences through comprehensive and robust text processing.

In Stage-I, we focus on utilizing the summarization capability of BART~\cite{lewis2019bart} to condense short high-relevant region descriptions into pseudo sentences (Figure~\ref{fig3}), capturing essential information from regions concisely. To begin, we retrieve local-level region descriptions from the Visual Genome (VG) dataset (a public dataset comprises region descriptions). However, since $47\%$ of VG images overlap with MSCOCO, we apply a duplicate-removal scheme~\cite{kuo2022beyond} to refine region descriptions. After annotating region descriptions, we utilize the pre-trained CLIP to retrieve proper region descriptions for each image. Given an image $I$, we apply the cosine similarity function to calculate the matching score for each region description, then rank these descriptions according to their scores in descending order, forming the ordered set of region descriptions $\hat{D}$. Subsequently, the top-$k$ most relevant descriptions are chosen based on their scores for the following steps, with the selection of $k$ detailed in Figure~\ref{fig6}. These selected top-$k$ region descriptions for the given image are denoted as $\hat{D}^{k}$. However, as illustrated in Figure~\ref{fig1} (b), the region descriptions lack modifying phrases typically found in standard sentences. Previous research, such as \citet{feng2019unsupervised}, indicates that concepts with minimal semantic content can lead to failures in image captioning training.

\begin{figure}[t]
\vspace{-0.0cm}
  \centering
  %\fbox{\rule{0pt}{2in} \rule{0.9\linewidth}{0pt}}
\includegraphics[width=0.9\linewidth]{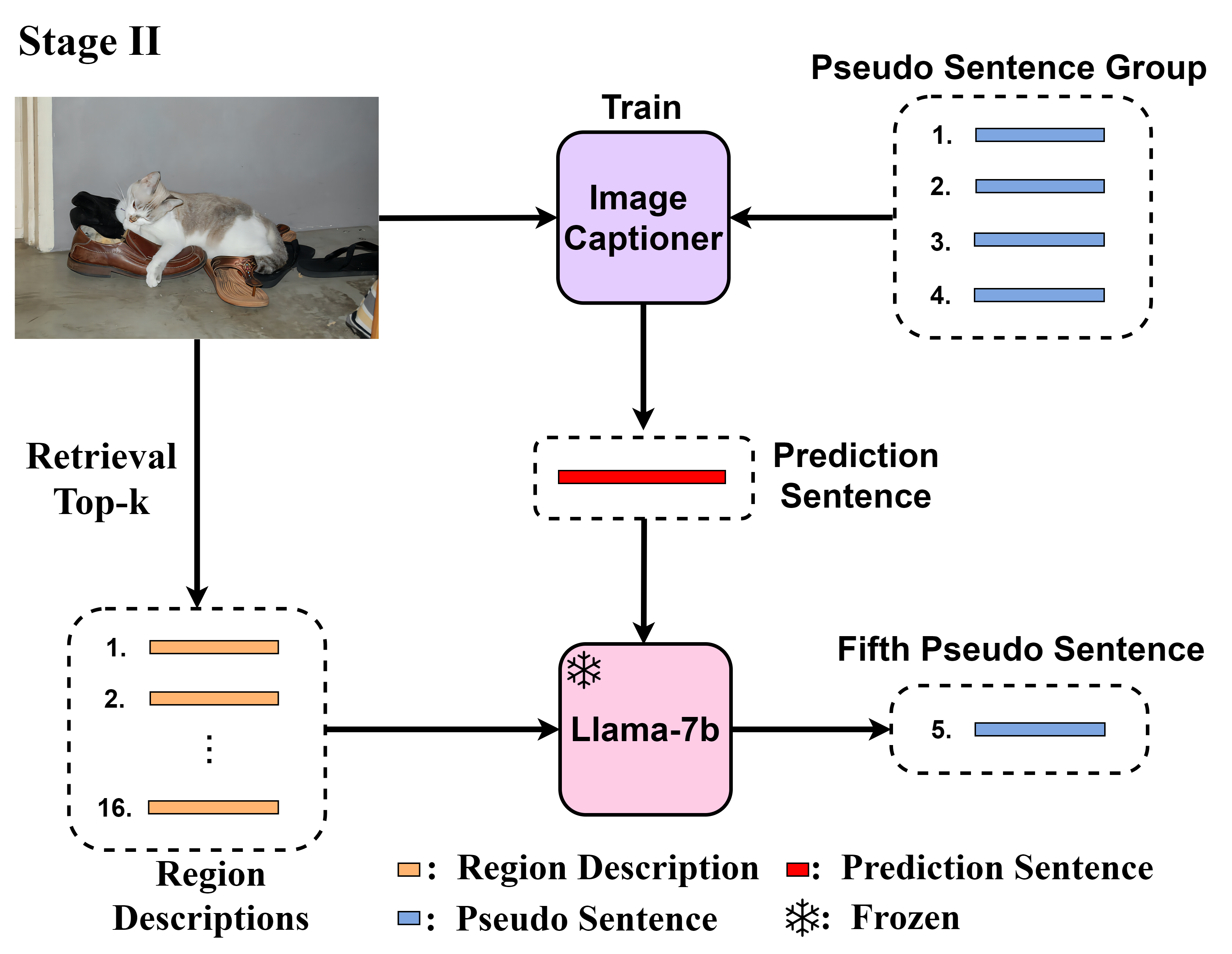}
\vspace{-0.0cm}
   \caption{The Stage-II of RaPSG framework. Initially, we utilize the provided image in conjunction with the preceding four pseudo sentences as supervision to train the image captioner. Once trained, we freeze the captioner and generate a prediction sentence. To enhance the generation process, we incorporate the top-$k$ most relevant region descriptions as supplementary material to get the fifth output.}
   \label{fig4}
   \vspace{-0.0cm}
\end{figure}

To cope with missing information, we refine local-level descriptions by summarizing them into global-level descriptions using BART. From the set $\hat{D}^{k}$, we select the top-$m$ descriptions with the criteria for choosing $m$ detailed in Figure~\ref{fig6}. Then, these descriptions are summarized into the first single sentence, $c_1$, by removing repeated words and leveraging the text summarization ability of BART (comparisons across different summarization models also depicted in Figure~\ref{fig6}). To enhance the diversity of pseudo sentences, instead of repeating the summarization process above, we group the remaining regions descriptions based on greater semantic differences. Specifically, a similarity score is calculated between each of the rest region descriptions $\hat{D}^{[k-m]}$ and the first pseudo sentence $c_1$. Next, these descriptions are grouped into $n$ comprehensive summarization sentences based on scores (i.e., $n = \frac{k-m}{m}$, the top $m$ for the $c_2$, the second top $m$ for the $c_3$, and ...). In this way, descriptions sharing more similarities would be grouped together to avoid arranging too many objects in a single sentence generation process. The issue of grouping complex objects together will be discussed in Section~\ref{fluency}. According to this setting, our method can generate a high-quality pseudo sentence group $\{c_i\}_{i=1}^{k/m}$ per image in the first stage.

In Stage-II, we distill crucial information from the preceding sentence group to generate more appropriate pseudo sentences. We refine the generated pseudo sentences in Stage-I using the expressive power of large generative models, producing fluent and contextually relevant sentences for supervision. Initially, we pair the set $\{c_i\}_{i=1}^{k/m}$ with the $I$ for captioner training, as shown in the top part of Figure~\ref{fig4}. This process establishes a reconnection between the sentences and the visual content, enabling the captioner's accuracy in both image and text domains. However, the supervision by pseudo sentences could lead the captioner to learn repeated information, potentially resulting in a lack of specific details within the context.

To address this limitation, we propose incorporating a large-size generative model, LLaMA-7B, to generate pseudo sentences with more detailed information. In our approach, we refine the sentences by using the predictions from the frozen captioner as well as the $\hat{D}^{k}$. By combining these elements, LLaMA learns the core ideas from the predictions and incorporates the detailed information from the region descriptions. This integration enables us to generate superior pseudo sentences that encompass a greater level of detail. Consequently, we obtain a more appropriate sentence as our another output denoting as $c_{k/m+1}$. With these two stages completed, we successfully generate a group of pseudo sentences $\{c_i\}^{k/m+1}_{i=1}$ that are ready for further use.

\subsection{Fluency Filter}
\label{fluency}
\begin{figure}[t]
  \centering
  %\fbox{\rule{0pt}{2in} \rule{0.9\linewidth}{0pt}}
   \includegraphics[width=0.92\linewidth]{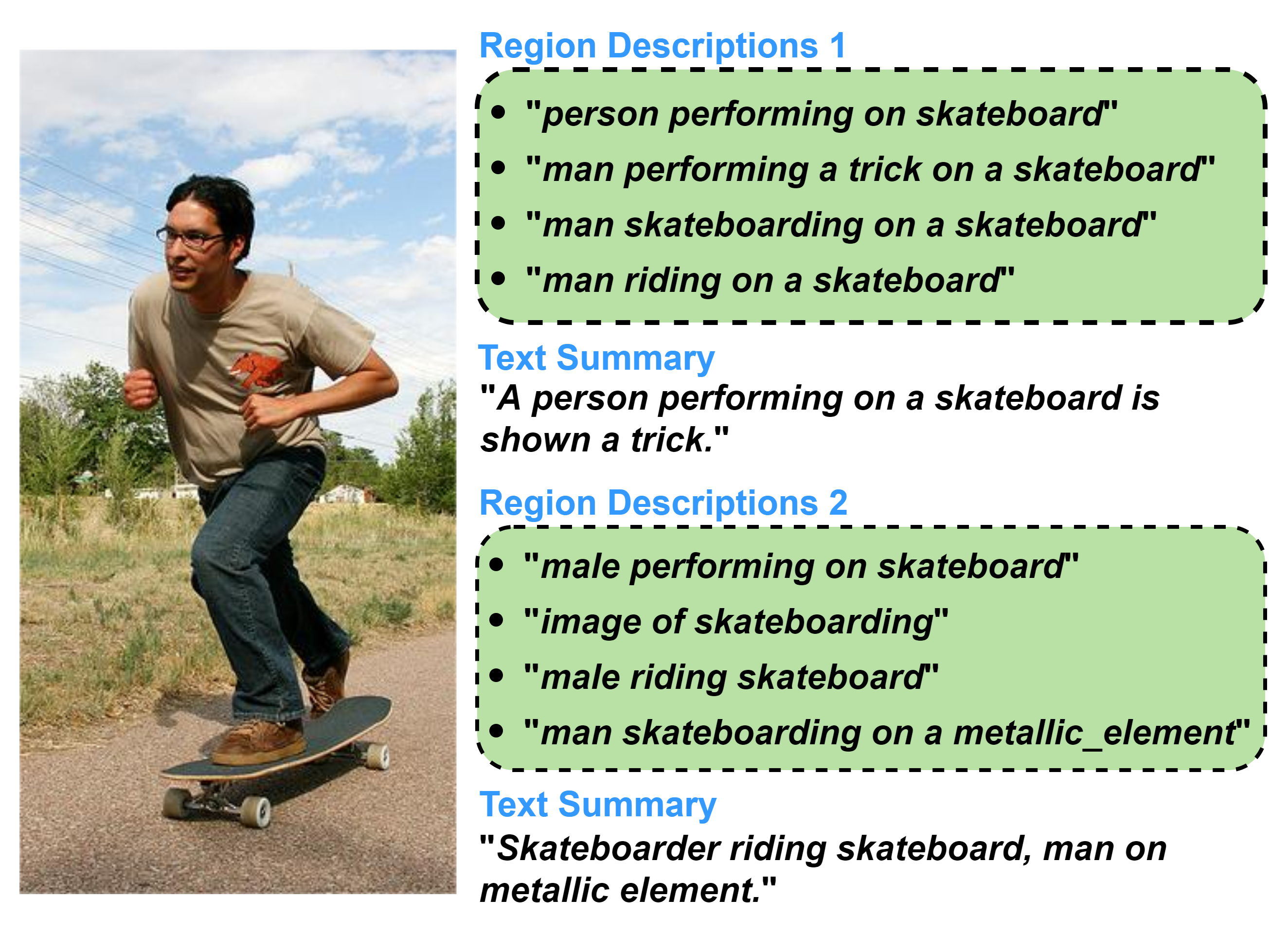}
   \vspace{-0.0cm}
   \caption{A comparison of two pseudo sentences in RaPSG process. The first sentence appears more fluent than the second sentence from the human view.}
   \label{fig5}
   \vspace{-0.0cm}
\end{figure}

The fluency filter is designed to sift the generated sentences to remove low-quality pseudo captions. For each given image $I$, the filter carefully selects the best sentence among $\{ c_i\}^{k/m+1}_{i=1}$ to ensure a precise match. Figure~\ref{fig5} compares two generated pseudo sentences from BART based on two groups of region descriptions in the first stage of the RaPSG module. The first case shows that the model successfully comprehends the relationship between the skateboard and the trick in the inference process. By contrast, the second sentence does not capture the important information to describe the image because the model recognizes the metallic-element different from the skateboard. Due to the limited discernment of LPMs, varying appellations for the same object in region descriptions can cause confusion, potentially fragmenting the generated sentence into multiple semantic parts and reducing its coherence and accuracy.

We propose to filter out the low-quality pseudo sentences via CIDEr metric~\cite{vedantam2015cider} (an image description evaluation based on human preference)
 because these low-quality pseudo sentences are also made up of highly relevant phrases but in an unnatural arrangement and can deceive the common evaluation methods. Since real annotations are unavailable, we use the model's predictions as references. To this end, we propose that the $\{ c_i\}^{k/m+1}_{i=1}$ are examined by the CIDEr metric, and the one graded the highest is chosen as follows:
 \begin{equation}
\begin{aligned}
    c_{cider} &= \underset{c}{\arg\max} CIDEr(c_i, f_c(I)),
  \label{eq3}
 \end{aligned}
\end{equation}
 where $c^j_i$ is the $j$-th pseudo sentence among five. $f_c(I_i)$ is the model prediction sentence and $f_c$ is the basic captioning model.  

\begin{table*}[t]
\caption{The comparison of our approach with SOTA zero-shot models on the Karpathy split of MSCOCO and Flickr30k benchmarks. We denote different captions (i.e., CTX, M$^2$, DIFNet, and DLCT) inside the brackets. Pseudo Sents. represents the generated pseudo sentences from the RaPSG module.}
\vspace{-0.0cm}
\fontsize{12}{16}\selectfont
\centering
 \scalebox{0.71}{
\begin{tabular}{lccc|cccccc|cccccc}
    \toprule[0.7mm]
    \multirow{2}{*}{\large  Model}  &Trainable & Pre-trained & External  & \multicolumn{6}{c|}{MSCOCO} & \multicolumn{6}{c}{Flickr30k} \\
      &Params & Models & Dataset & B1 & B4 & M & C & S & CLIP-S & B1 & B4 & M & C & S & CLIP-S\\
    \midrule[0.3mm]
     SimVLM$_{base}$~\shortcite{wang2021simvlm} & - & \multirow{2}{*}{-} & \multirow{2}{*}{1.8B} & - & 9.5 & 11.5 & 24.0 & 7.5 & - & - & - & - & - & - & - \\
    SimVLM$_{huge}$~\shortcite{wang2021simvlm} &1.4B &  &  & - & 11.2 & 14.7 & 32.2 & 8.5 & - & - & - & - & - & - & -\\
    \midrule[0.3mm]
    Flamingo3B~\shortcite{alayrac2022flamingo} & 1.3B & NFNet~\shortcite{brock2021high} and & \multirow{2}{*}{312M} & -  & -& - & 73.0 & - & - & - & - & -& 60.6 & - & -\\
    Flamingo80B~\shortcite{alayrac2022flamingo} & 10B & Chinchilla~\shortcite{hoffmann2022training} &  & - &  - &  - & \textbf{84.3} & - & - & - & - & - & \textbf{67.2} & - & -\\
    \midrule[0.3mm]
    MiniGPT4-V1~\shortcite{zhu2023minigpt} & - & ViT~\shortcite{dosovitskiy2020image} and & 5M & 23.6 & 5.8 & 20.9 & 0.0 & 14.4 & 34.0 & 13.2 & 3.5 & 15.6 & 0.0 & 14.8 & 32.3\\
    MiniGPT4-V2~\shortcite{chen2023minigpt} & - & Vicuna~\shortcite{chiang2023vicuna} & 20M & 28.6 & 6.3 & 24.4 & 0.0 & 17.9 & 35.5 & 17.5 & 6.6 & 22.0 & 0.0 & 20.0 & 32.6\\
    \midrule[0.3mm]
    LLaVA1.0~\shortcite{liu2023improved} & 0.14B & CLIP~\shortcite{radford2021learning} and & 0.59M & 38.5 & 9.1 & \textbf{26.7} & 50.9  & \textbf{24.2} & 34.1 & 48.0 & 13.0 &  \textbf{23.4} & 52.5 & \textbf{17.1} & 33.7\\
    LLaVA1.5~\shortcite{liu2024visual} & 0.70B & Vicuna~\shortcite{chiang2023vicuna}  & 0.66M & 30.6 & 10.1 & 24.8 & 41.8 & 22.6 & 31.5 & 35.2 & 7.7 & 21.9 & 34.1 & 17.0 & 30.5\\
    \midrule[0.3mm]
    Our Pseudo Sents. & 0 & \multirow{2}{*}{CLIP~\shortcite{radford2021learning},} & \multirow{5}{*}{\textbf{0.45M}} & 48.1 & 8.8 & 18.0 & 39.3 & 13.3 & \textbf{47.6} & 43.2 & 14.5 & 17.1 & 21.2 & 9.3 & \textbf{45.4}\\
    Ours (w/ CTX) & 40M & & & 67.0 & 18.3 & 21.2  & 72.4 & 14.1 & 33.6 & 51.7 & 17.8 & 21.0 & 53.3 & 10.7 & 32.6\\
    Ours (w/ M$^2$) & 38M & BART~\shortcite{lewis2019bart}, and & & 67.5 & 18.9 & 20.9  & 75.3 & 14.7 & 34.3 & 54.6 & 17.5 & 20.7 & 56.8 & 11.2 & 33.8\\
    Ours (w/ DLCT) & 63M & \multirow{2}{*}{LLaMA~\shortcite{touvron2023llama}} &  & 69.5 & \textbf{19.4} & 21.1  & 75.9 & 14.5  & 34.5 & 54.1 & 18.1 & 22.6 & 58.4 & 11.5 & 34.1\\
    Ours (w/ DIFNet) & \textbf{33M} & & & \textbf{70.5}  & 19.3 & 21.4  & 78.1 & 14.9 & 35.8 & \textbf{55.9} &  \textbf{18.2} & 23.1 & 59.1 & 11.8 & 33.9\\
    \bottomrule[0.7mm]
\end{tabular}}
\label{tab1}
\vspace{-0.0cm}
\end{table*}
 
\subsection{CLIP Guidance}
% \Chaoyi{The naming might be not technically correct. It's an additional training loss/object, rather than a module. Maybe change "Clip-based reward" to "CLIP Guidance"? Does it sound good to you? Please review.}
\label{contrastive}
The CLIP guidance module is proposed to encourage the sentence prediction to semantically match image content in CLIP embedding space as we abandon pre-training on external large-scale datasets. The InfoNCE~\cite{oord2018representation} is employed to reduce cross-modal information loss. The frozen image encoder CLIP-I and text encoder CLIP-T are used to embed a dozen original images and corresponding predictions into a shared semantic space. Then, the pairwise affinities are computed based on the encoded features. The learning process can be formulated as minimizing the contrastive information loss:
\begin{equation}
\begin{aligned}
    &L_I =-\text{log}\frac{\text{exp}(q\cdot k^+ / \tau)}{\text{exp}(q\cdot k^+ / \tau)+\sum\limits_{k^-}\text{exp}(q\cdot k^- / \tau)},
  \label{eq5}
 \end{aligned}
\end{equation}

\noindent where $q$ is a visual embedding for an image extracted from the CLIP-I, $k^+$ is the text embedding for this image (positive key), and $k^-$ are text embedding for other images from the same batch in the training process (negative key). Both of them are generated by CLIP-T. $\tau$ is the temperature hyper-parameter.

\section{Experiments and Results}
\subsection{Experiments Setting}
\textbf{Datasets.}We choose MSCOCO~\cite{chen2015microsoft} and Flickr30k~\cite{plummer2015flickr30k} with Karpathy~\cite{karpathy2015deep} split as our test benchmark. The MSCOCO images are divided into three parts: 113k images for training, 5k images for validation, and the remaining 5k images for testing. The Flickr30k images are divided into three parts: 29k images for training, 1k images for validation, and the remaining 1k images for testing.

\textbf{Evaluation Metrics.}Following standard captioning evaluation protocols~\cite{li2019entangled}, we employ the following five metrics:  BLEU~\cite{papineni2002bleu}, METEOR~\cite{banerjee2005meteor}, ROUGE~\cite{lin2004rouge}, CIDEr~\cite{vedantam2015cider}, and SPICE~\cite{anderson2016spice}. Beyond these traditional metrics, we also incorporate the innovative robust metric CLIP-S~\cite{hessel2021clipscore}, which assesses the relevance between the generated caption and the target image independently of reference captions. 

\textbf{Image Captioning Backbones.} Our approach is versatile for different image captioning models. To validate its performance, we incorporate our proposed framework with several classic captioners, including: M$^2$ model~\cite{cornia2020meshed}, CTX model~\cite{kuo2022beyond}, DLCT model~\cite{luo2021dual}, and DIFNet model~\cite{wu2022difnet}.

\subsection{Comparison against Large Pre-Trained Models}
We compare our RaPSG approach with the zero-shot models~\cite{wang2021simvlm,alayrac2022flamingo,zhu2023minigpt,liu2024visual} on MSCOCO and Flickr30k benchmarks, as they are all built up on LPMs. Table~\ref{tab1} demonstrates that our method surpasses the performance of these models on the MSCOCO benchmark in some metrics  (Note that multimodal LLMs like MiniGPT4 and LLaVA are not specifically trained to generate short captions, and their detailed descriptions may not be fully captured by traditional metrics). Moreover, previous approaches rely on pre-training with a large number of external image-text pairs and demand a considerable number of trainable parameters. For instance, Flamingo3B is pre-trained on 312M external image-text pairs, whereas our model only requires 0.45M (0.14\%) generated pseudo sentences, which is more data-efficient. Additionally, we also validate our approach on another popular benchmark Flickr30k. Table~\ref{tab1} shows our method's robustness across datasets, matching SOTA models in performance with fewer trainable parameters (e.g., $6.7\%$ of Flamingo).

\subsection{Comparsion against Finetuning-Based Approaches}
Next, we compare ours with other models that operate without full supervision, including unsupervised~\cite{zhou2021triple,honda2021removing}, unpaired~\cite{ben2021unpaired,liu2021exploring}, and weakly-supervised~\cite{zhang2022look,zhu2022unpaired} approaches. Unsupervised and weakly-supervised methods retrieve sentences from mismatching corpora, while unpaired methods use the original corpora but each sentence does not pair with the corresponding images.
Table~\ref{tab2} indicates that our method surpasses these data-efficient methods by utilizing the generated pseudo sentences instead of fetching complete sentences. It is significant to note that our method even surpasses unpaired setting models that employ real images and real annotations but operate in an unpaired setting. This suggests that generating pseudo sentences may hold greater potential than retrieving sentences. 

\begin{table}[t]
  \centering
  \caption{The comparison of our method and others without fully supervision on MSCOCO benchmark.}
  \vspace{-0.0cm}
  \scalebox{0.8}{
  \begin{tabular}{llcccccc}
    \toprule[0.5mm]
    Category & Method & B1 & B4 & M & R & C & S\\
    \midrule[0.3mm]
    \multirow{2}{*}{\shortstack[l]{Unsupervised\\ Setting}}
    &TSGAN~\shortcite{zhou2021triple} & 46.2 & 6.9 & 13.0 & 32.3 & 28.9 & 8.3\\
    &RWLSA~\shortcite{honda2021removing} & 50.2 & 6.8 & 14.1 & 34.8 & 32.9 & 8.8\\
    \midrule[0.3mm]
    \multirow{2}{*}{\shortstack[l]{Unpaired\\ Setting}}
    &SCS~\shortcite{ben2021unpaired} & 67.1 & \textbf{22.8} &21.4 & 47.7 &74.7 &15.1\\
    &FG-SRE~\shortcite{liu2021exploring} & 67.8 & 21.8 & \textbf{22.1} & 48.4 & 75.7 & \textbf{16.1}\\
    \midrule[0.3mm]
    \multirow{2}{*}{\shortstack[l]{Weakly-super-\\ vised Setting}}
    &SGCL~\shortcite{zhang2022look} &63.6 & 20.2 & 20.0 & 47.9 & 55.0 & 13.5\\
    &WS-UIC~\shortcite{zhu2022unpaired}  & - & 21.5 & 20.1 & 45.8 & 65.7 & 13.6\\
    \midrule[0.3mm]
    \multirow{1}{*}{\shortstack[l]{LPMs + RaPSG}}
    & Ours (w/ DIFNet) & \textbf{70.5} & 19.3 & 21.4 & \textbf{49.0} & \textbf{78.1} & 14.9\\
    \bottomrule[0.5mm]
  \end{tabular}}
     \label{tab2}
     \vspace{-0.0cm}
\end{table}

\begin{table}[t]
 \caption{Performance comparison with SOTA cross-domain methods on MSCOCO and Flickr30k captioning tasks.}
 \vspace{-0.0cm}
 \scalebox{0.84}{
\begin{tabular}{llccccc}
    \toprule[0.5mm]
Direction                          & Method          & B4   & M    & R    & C    & S    \\ \hline
\multirow{4}{*}{COCO-to-Flickr30k} & DeCap~\cite{li2023decap}           & 16.3 & 17.9 & -    & 35.7 & 11.1 \\
                                   & CapDec~\cite{nukrai2022text}          & 17.3 & 18.6 & 42.7 & 35.7 & -    \\
                                   & CgT-GAN~\cite{yu2023cgt}         & 17.3 & 19.3 & 43.9 & 47.5 & \textbf{12.9} \\
                                   & Ours (w/DIFNet) & \textbf{17.1} & \textbf{20.2} & \textbf{44.6} & \textbf{51.3} & 11.6 \\ \hline
\multirow{4}{*}{Flickr30k-to-COCO} & DeCap~\cite{li2023decap}           & 9.2  & 16.3 & 36.7 & 27.3 & -    \\
                                   & CapDec~\cite{nukrai2022text}          & 12.1 & 18.0 & -    & 44.4 & 10.9 \\
                                   & CgT-GAN~\cite{yu2023cgt}         & 15.2 & 19.4 & 40.9 & 58.7 & \textbf{13.4} \\
                                   & Ours (w/DIFNet) & \textbf{17.7} & \textbf{20.1} & \textbf{45.7} & \textbf{66.3} & 12.2 \\ \bottomrule[0.5mm]
\end{tabular}}
\label{tab3}
\vspace{-0.0cm}
\end{table}

\begin{figure*}[t]
  \centering
  %\fbox{\rule{0pt}{2in} \rule{0.9\linewidth}{0pt}}
   \includegraphics[width=0.99\linewidth]{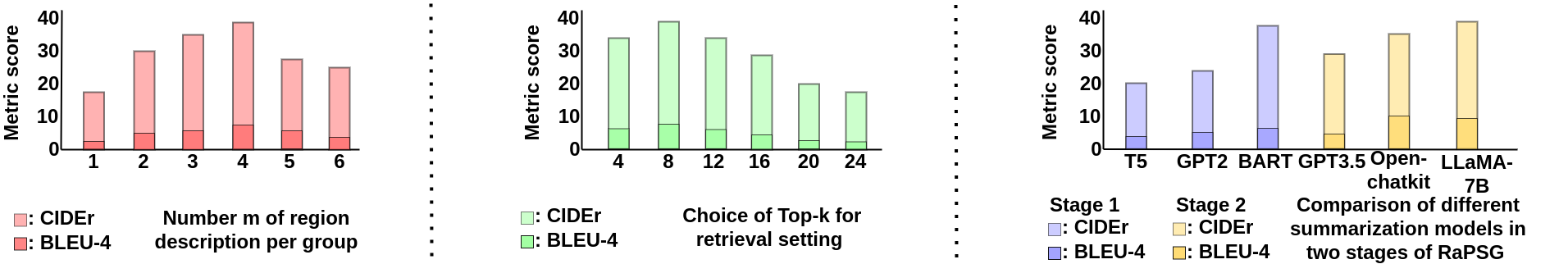}
   \vspace{-0.0cm}
   \caption{We conduct experiments to compare different settings, aiming to determine the most effective method for generating high-quality pseudo sentences based on region descriptions. These comparisons include, from left to right: the selection of hyperparameter $m$, the choice of hyperparameter $k$, and the evaluation of various summarization models.}
   \label{fig6}
   \vspace{-0.0cm}
\end{figure*}

\subsection{Extension on Cross-Domain Image Captioning Benchmarks}
To further verify the robustness of our model, we evaluate it on a cross-domain image captioning benchmark in comparison with SOTA models~\cite{li2023decap,nukrai2022text,yu2023cgt}. Notably, we adhered to the established cross-domain image captioning benchmark protocol~\cite{laina2019towards}, albeit with the textual corpora replaced by the VG dataset. As detailed in Table~\ref{tab3}, our model also demonstrates a significant improvement, with CIDEr scores of 51.3 (+3.8) and 66.3 (+7.6) compared to competing models in two assessed categories.

\subsection{Extension on Semi-Supervised Image Captioning Benchmarks}
We also test whether our approach can deal with the data scarcity problem in a semi-supervised setting where only partial images have the corresponding text annotations. Specifically, we follow the existing semi-supervised image captioning benchmark~\cite{chen2021self}. The proposed RaPSG is firstly optimized on the $99\%$ images without caption labels. Then, the model is further finetuned on the rest of $1\%$ labelled data. We repeat the experiments under 3 times and calculate the average performance as output. As shown in Table~\ref{tab4}, compared with current approaches, our approach achieves a performance gain with a 93.4 (+8.9) CIDEr score.

\begin{figure*}[t]
  \centering
  %\fbox{\rule{0pt}{2in} \rule{0.9\linewidth}{0pt}}
   \includegraphics[width=0.99
   \linewidth]{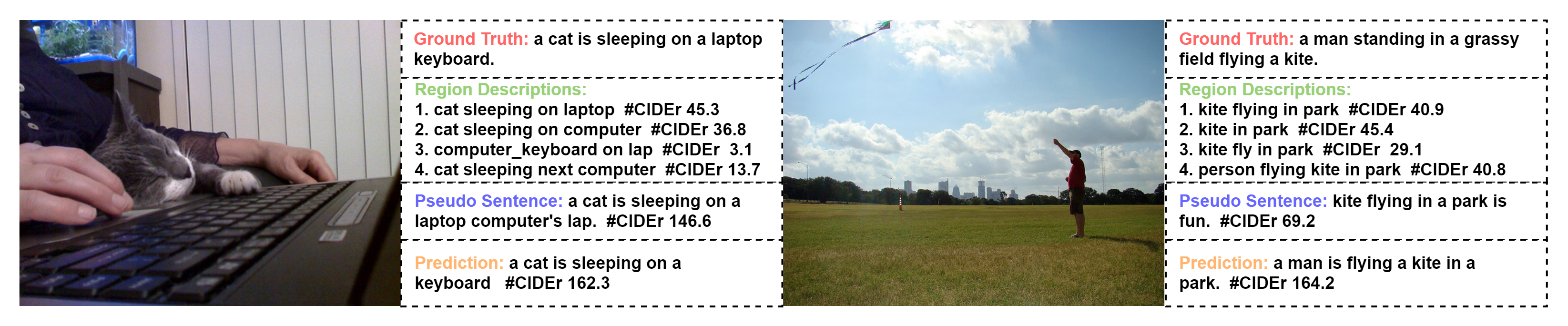}
   \vspace{-0.0cm}
   \caption{Qualitative results of our approach based on DIFNet model including region descriptions, pseudo sentences, and final predictions. Best viewed by zooming in.}
   \label{fig7}
   \vspace{-0.0cm}
\end{figure*}

\subsection{Ablation Studies}
\begin{table}[t]
\centering
 \caption{Performance comparison with SOTA LPM-based method on $1\%$ semi-supervised MSCOCO captioning task.}
 \vspace{-0.0cm}
\scalebox{0.95}{
  \begin{tabular}{lcccccccccc}
    \toprule[0.5mm]
    Model & B1  & B4 & M & R & C & S  \\
    \midrule[0.3mm]
    Self Distillation~\shortcite{chen2021self}  &67.9  & 25.0 & 21.7 & 49.3 & 73.0 & 14.5\\
    OSCAR~\shortcite{li2020oscar}  & 67.2  & 23.3 & 22.5 & 49.1 & 78.4 & -\\
    VisualGPT~\shortcite{chen2022visualgpt}  & 69.5  & 25.6 & 22.6 & 49.6 & 80.9 & -\\
    P$^3$~\shortcite{jain2021perturb} &68.8  & 27.5 & \textbf{23.4} & 51.0 & 84.5 & 16.1\\
    \midrule[0.3mm]
    %M$^2$~\shortcite{cornia2020meshed} &67.4 &22.8 &21.4 &48.2 & 70.9 &14.8\\
    %DIFNet~\shortcite{wu2022difnet} & 70.8  & 24.9 & 22.2 & 49.7 & 81.3 & 15.3 \\
    \textcolor{gray}{DIFNet~\shortcite{wu2022difnet}} & \textcolor{gray}{70.8}  & \textcolor{gray}{24.9} & \textcolor{gray}{22.2} & \textcolor{gray}{49.7} & \textcolor{gray}{81.3} & \textcolor{gray}{15.3} \\
    \textbf{DIFNet + Ours} & 7\textbf{3.5}  & \textbf{27.7}  & 23.1 & \textbf{51.8} & \textbf{93.4} & 16.7 \\
    \bottomrule[0.5mm]
  \end{tabular}}
  \label{tab4}
  \vspace{-0.0cm}
\end{table}

\begin{table}[t]
  \centering
  \caption{Ablation study of different proposed modules conducted on the DIFNet. "RD" represents the retrieved region descriptions. "PS" means the generated pseudo sentences. "FF" is the fluency filter. "CG" represents the CLIP guidance. "D" is the DIFNet model.}
  \vspace{-0.0cm}
  \scalebox{0.88}{
  \begin{tabular}{lcccccccccc}
    \toprule[0.4mm]
    Module & B1 & B4 & M & R & C & S & CLIP-S\\
    \midrule[0.3mm]
    RD &10.6 & 1.5 & 9.8 & 19.3 & 18.3 &7.5 & \textbf{62.7}\\
    PS & 48.1 & 8.8 &18.0 & 33.8 & 39.3 & 13.3 & 47.6\\
    PS+FF & 59.4 & 15.2 & 20.2 & 39.6 & 56.0 & 13.9 & 48.2\\
    D+PS & 67.9 & 16.9 & 20.3 & 43.9 & 70.2 & 13.7 & 31.5\\
    D+PS+FF & 70.3 & 19.1 & 21.1 & 45.9 & 76.9 & 14.7 & 32.1\\
    D+PS+FF+CG & \textbf{70.5} & \textbf{19.3} & \textbf{21.4} & \textbf{46.0} & \textbf{78.1} & \textbf{14.9} & 35.8\\
    \bottomrule[0.4mm]
  \end{tabular}}
  \label{tab5}
  \vspace{-0.0cm}
\end{table}

\textbf{Contribution of Designed Modules.} We investigate the contribution of each designed module, as shown in Table~\ref{tab5}. The RaPSG module is the key component to improving the model performance. In addition, the fluency filter is designed to filter out the unnatural sentences among pseudo sentence generation and leave the best one matching the given image. Finally, we introduce CLIP guidance in the retrieval-augmented learning process, which drives the prediction to be semantically consistent with the given image by shrinking the cross-modal distance in the feature embedding space.

\textbf{Pseudo Sentence Quality.}
Here, we explore how to regulate the quality of generated sentences and the methods for producing high-quality sentences. We first investigate the parameter $m$ to ascertain the generation of high-quality pseudo sentences. Subsequently, based on the chosen value of $m$, we explore the selection of top-$k$. According to the left part of Figure~\ref{fig6}, we decide to set $m=4$ as it yields the best performance. Then, based on the $m$ value, we explore $k \in [4, 24]$. As suggested by the middle segment of Figure~\ref{fig6}, we set $k = 16$ and disregard the subsequent region descriptions according to the CIDEr scores. Lastly, we evaluate the efficacy of various summarization models. Based on the result in the right section of Figure~\ref{fig6}, we select the BART for the initial stage and the LLaMA-7B for the subsequent phase of RaPSG.

\subsection{Qualitative Results} 
To highlight the ability of our approach, we present some qualitative results of our generated pseudo sentences and predictions in Figure~\ref{fig7}. It can be noticed that our predictions have better quality under all metrics compared with region descriptions and pseudo sentences. However, some of the examples do not make sense from a human view though their CIDEr scores are fine (e.g., ``\textit{kite flying in a park is fun.}"). The correct sentence would be like ``\textit{flying a kite in a park is fun.}". One potential reason could be the pre-trained text summarization BART cannot deal with a batch of similar objects in the composition process. It tends to disarrange complex relationships among different region descriptions.

\section{Conclusion}
In this work, we propose a retrieval-augmented pseudo sentence generation method which leverages the prior knowledge from the frozen LPMs. The generated sentences can avoid the appearance of irrelevant words and keep the diversity of pseudo references, which is attributed to the innovative combination of ranking, grouping, and summarization. In addition, we design a fluency filter to sift the generated sentences and a CLIP guidance module to make the predicted captions semantically consistent with the given image. Our approach outperforms existing state-of-the-art captioning models across various scenarios such as zero-shot, unsupervised, semi-supervised, and cross-domain settings.

%%
%% The next two lines define the bibliography style to be used, and
%% the bibliography file.
\bibliographystyle{ACM-Reference-Format}
\bibliography{sample-base}

%%
%% If your work has an appendix, this is the place to put it.
% \appendix

% \section{Research Methods}

% \subsection{Part One}

% Lorem ipsum dolor sit amet, consectetur adipiscing elit. Morbi
% malesuada, quam in pulvinar varius, metus nunc fermentum urna, id
% sollicitudin purus odio sit amet enim. Aliquam ullamcorper eu ipsum
% vel mollis. Curabitur quis dictum nisl. Phasellus vel semper risus, et
% lacinia dolor. Integer ultricies commodo sem nec semper.

% \subsection{Part Two}

% Etiam commodo feugiat nisl pulvinar pellentesque. Etiam auctor sodales
% ligula, non varius nibh pulvinar semper. Suspendisse nec lectus non
% ipsum convallis congue hendrerit vitae sapien. Donec at laoreet
% eros. Vivamus non purus placerat, scelerisque diam eu, cursus
% ante. Etiam aliquam tortor auctor efficitur mattis.

% \section{Online Resources}

% Nam id fermentum dui. Suspendisse sagittis tortor a nulla mollis, in
% pulvinar ex pretium. Sed interdum orci quis metus euismod, et sagittis
% enim maximus. Vestibulum gravida massa ut felis suscipit
% congue. Quisque mattis elit a risus ultrices commodo venenatis eget
% dui. Etiam sagittis eleifend elementum.

% Nam interdum magna at lectus dignissim, ac dignissim lorem
% rhoncus. Maecenas eu arcu ac neque placerat aliquam. Nunc pulvinar
% massa et mattis lacinia.

\end{document}